\documentclass[11pt,twoside]{article}
\usepackage{fullpage}
\usepackage{ifthen}
\usepackage{epsf}
\usepackage{fancyheadings}
\usepackage{hyperref}
\usepackage{url}
\usepackage{enumerate}

 \usepackage{amsthm}
 \usepackage{amsfonts}
 \usepackage{amsmath}
 \usepackage{amssymb}
 \usepackage{color}
\usepackage{microtype}
\usepackage{multicol}
 \usepackage{amsfonts}
 \usepackage{amsmath}
 \usepackage{amssymb}

\usepackage{graphics}
\usepackage{graphicx}
\usepackage{mathrsfs}
\usepackage{float}

\usepackage{epsf}
 \usepackage{subfigure}
 \usepackage{caption}
 
\newlength{\widebarargwidth}
\newlength{\widebarargheight}
\newlength{\widebarargdepth}

\makeatletter
\long\def\@makecaption#1#2{
        \vskip 0.8ex
        \setbox\@tempboxa\hbox{\small {\bf #1:} #2}
        \parindent 1.5em 
        \dimen0=\hsize
        \advance\dimen0 by -3em
        \ifdim \wd\@tempboxa >\dimen0
                \hbox to \hsize{
                        \parindent 0em
                        \hfil 
                        \parbox{\dimen0}{\def\baselinestretch{0.96}\small
                                {\bf #1.} #2
                                } 
                        \hfil}
        \else \hbox to \hsize{\hfil \box\@tempboxa \hfil}
        \fi
        }
\makeatother

\newtheorem{theorem}{Theorem}

\usepackage{algorithmic}
\usepackage[ruled,lined,english]{algorithm2e}

\begin{document}

\title{Reinforcement Learning algorithms for regret minimization in structured Markov Decision Processes}
\author{Prabuchandran K J$^{1}$, Tejas Bodas$^{2}$ and Theja Tulabandhula$^{3}$\\
$^{1}$Indian Institute of Science, India\\
$^{2}$Indian Institute of Technology Bombay, India\\
$^{3}$Xerox Research Centre India}
\date{January 30, 2016}
\maketitle

\begin{abstract}
A recent goal in the Reinforcement Learning (RL) framework is to choose a sequence of actions or a policy to maximize the reward collected or minimize the regret incurred in \emph{a finite time horizon}. For several RL problems in operation research and optimal control, the optimal policy of the underlying Markov Decision Process (MDP) is characterized by a known structure. The current state of the art algorithms do not utilize this known structure of the optimal policy while minimizing regret. In this work, we develop new RL algorithms that exploit the structure of the optimal policy to minimize regret. Numerical experiments on MDPs with structured optimal policies show that our algorithms have better performance, are easy to implement, have a smaller run-time and require less number of random number generations.

\end{abstract}

\section{Introduction}
\label{introduction}

\noindent Optimal sequential decision problems are common in artificial intelligence (chess, Go, Backgammon), operation research (inventory 
control management, queueing admission control) and various other scientific disciplines. These problems are modeled using the framework of MDPs where a goal is to find a policy that maximizes the long run average reward collected. In an important subclass of these problems, one can characterize the structure of the optimal policy that maximizes this long run average reward. For example, the multi-period $(s,S)$ policy in inventory control management \cite{Bertsekas95,Puterman14}, the threshold and multi-threshold policies in queueing systems \cite{Duff95,Krishnamurthy07,Yang13,Lin84,Gogate99} and communication systems \cite{Altman10} and the threshold policy in machine replacement problems \cite{Love00} are MDP problems where the optimal policy is characterized by a special structure.  For such class of problems, one can develop efficient algorithms to obtain the optimal policy by restricting the search over such structured policies.
   
In the MDP problems outlined above, the modeling parameters such as the transition probability and the reward/cost matrices are assumed to be known. However in many practical scenarios, these parameters are unknown. This is the realm of RL where the objective is to learn the optimal policy online based on the simulation sample. RL algorithms such as Q-learning \cite{Watkins92} and SARSA \cite{Sutton98} achieve this objective. However these algorithms learn the optimal policy only asymptotically making them unsuitable for many practical problems. This has led to interest in developing algorithms that have finite time guarantees. In these algorithms the aim is to bound the number of  samples required for finding an $\epsilon$-optimal policy  with high probability.  EXP3 \cite{Kearns02} and Rmax \cite{Brafman03} algorithms provide such PAC guarantees while learning the optimal behaviour. 
More recently the focus in RL problems is to maximize the cumulative reward collected (or equivalently minimize the total regret). Algorithms that pursue regret minimization in a finite time horizon are the UCRL \cite{Ortner07,Jaksch10}, PSRL \cite{Osband13}, Thompson PSRL \cite{Gopalan15} and the RLPA \cite{azar2013regret} algorithms. 
These algorithms emphasize exploration over unexplored policies, but also ensure that the best policy discovered so far is used frequently. 

In this paper, our aim is to minimize the cumulative regret in a finite time horizon for the class of RL problems where the structure of the optimal policy is known. We propose algorithms that utilize this structural information to restrict the policy search. Note that none of the existing algorithms exploit the known structure of the optimal policy when minimizing the cumulative regret. Further, in case of UCRL \cite{Jaksch10} and  PSRL \cite{Osband13}, certain explored policies chosen by the algorithm need not satisfy the structural property of the optimal policy. 

As is the case with UCRL and PSRL, our regret minimization algorithms are also inspired from the regret minimization algorithms of the multi-arm bandit (MAB) problem (\cite{Auer02}, \cite{Agrawal11}). 
We provide three algorithms, \emph{pUCB, pThompson and warmPSRL}, that treat `structured policies' as arms of a corresponding MAB problem.
As the names suggest, our algorithms are conceptually inspired by the UCB \cite{Auer02} and Thompson sampling \cite{Agrawal11} algorithms for the MAB problem. 
While the UCRL algorithm is also related to the UCB algorithm, our pUCB algorithm is very different from UCRL. The UCRL algorithm maintains upper confidence estimates on the entries of the transition probability and the reward matrix. On the other hand, pUCB maintains such confidence bounds on the estimate for the average reward of a policy. The state of the art PSRL algorithm is based on Thompson sampling \cite{Agrawal11} and maintains a belief on the transition probability and the reward matrix. In contrast, our pThompson algorithm maintains a belief on the average rewards of policies. This makes our algorithms model-free in contrast to UCRL and PSRL. 

Our pUCB and pThompson algorithms  are  episodic in nature, i.e., a chosen policy is implemented for a number of rounds before being changed. Now, to be able to estimate the average reward of a policy correctly we require unbiased estimates for the same.  To ensure this, we  use the notion of recurrence times and implement the chosen policy for an episode where the starting state and the end state are the same. 
This is a nontrivial technicality that some of the previous algorithms such as the RLPA \cite{azar2013regret} overlooked by assuming episodes of arbitrary random duration.

Our idea of treating policies as arms and converting a RL problem into a corresponding MAB problem cannot be used in general for any RL problem.  For an MDP with $M$ actions and $N$ states, there are potentially $M^N$ policies. Treating all such policies as arms and then applying either a UCB-like or a Thompson-like algorithm may not be productive in minimizing regret.  However as we show with our numerical experiments, this is not the case when the structure of the optimal policy is utilized. Our pUCB and pThompson algorithms turn out to be meaningful for the class of sequential decision problems with a structure on the optimal policy.

The rest of the paper is organized as follows. In the next section, we explain the requisite preliminaries for describing our algorithms. In Section \ref{sec:palgos}, we describe our pUCB, pThompson and warmPSRL algorithms. In Section \ref{sec:numerical}, we provide two MDP settings where the optimal policy has a known structure. We perform numerical experiments in both settings to analyze the performance of our algorithms. We conclude in Section \ref{sec:summary} with a brief summary and some future direction.

\section{Preliminaries and Motivation}
\label{sec:prelim}
\noindent A finite MDP consists of state space $S=\{1,2,\ldots,N\}$, action space $A=\{1,2,\ldots,M\}$, a probability transition matrix $P = [[p_{i,j}(a)]]$ where $ i,j \in S $ and $ a \in A.$ Here $p_{i,j}(a)$  denotes the probability of a transition to state $j$ when action $a$ is chosen at state $i$. The reward is characterized by the function $R(i,a,j)$ $\forall i,j \in S, a \in A $ that specifies the reward for choosing action $a$ in state $i$ and transitioning to $j$.  We assume the reward function is bounded by $1$. A stationary deterministic policy $\pi$ specifies the action ($\pi(i)$) that needs to be chosen in the state $i$. Let $\Pi$ denote the set of all deterministic stationary policies. 
Associated with each policy $\pi$, we define the long-run average reward $\rho(\pi)$ as ${\displaystyle
\rho(\pi) = \lim_{T \rightarrow \infty}\frac{\mathbb{E}\left[\sum_{t=1}^T R(s_t,\pi(s_t),s_{t+1})\right]}{T}},$
where we have suppressed the dependence on the starting state in the notation. The classical goal in the MDP framework is to find a policy $\pi^*$ that maximizes the long-term average reward, i.e.,
\begin{align*}
\pi^* = \arg \max_{\pi \in \Pi} \rho(\pi).
\end{align*}

This problem is well behaved if $\rho(\pi)$ is the same for all starting states. A sufficient condition for this to happen (and also assumed in this paper) is that the MDP satisfies the \textit{unichain condition} \cite{tsitsiklis2007np}. That is, for every policy $\pi$, the resulting Markov chain has a single ergodic class.

In the RL setting that is of interest to us, the transition probabilities and reward values are not known a priori, thus making it harder to compute $\pi^*$. In many modern RL settings, an alternative goal involves finding a sequence of policy $\{\pi_t\}$ that minimizes the expected \emph{regret} in a given horizon $T$. Here, the expected regret $\mathcal{R}_{s_{start}}(T)$ when started from state $s_{start}$ is defined as:
\begin{align*}
\mathcal{R}_{s_{start}}(T)= \rho(\pi^*)T -\overset{T}{\underset{t=1}{\sum}}\mathbb{E}[R(s_t,\pi_t(s_t))],
\end{align*}
where $\pi_{t}$ represents the policy chosen at time $t$.
  
In this paper we are concerned with minimizing regret when there exists some structure in the optimal policy $\pi^*$. Such structure exists in a wide collection of MDPs including instances in inventory management, optimal control (for instance, linear-quadratic regulators), queueing systems, dynamic resource allocation problems and others. We discuss two such MDP settings in Section \ref{sec:numerical}. Now using the structure of the optimal policy, one can create a smaller subset of policies that satisfies this structure. The size of this set can be polynomial in the number of states as opposed to exponential (this is problem dependent). We can search for a policy that minimizes regret using this set as compared to searching in the complete policy space. Our algorithms are designed to take advantage of the structure present in the optimal policy of the unknown MDP.  To motivate these, we give an interpretation of the PSRL algorithm that naturally leads to our algorithms that use the structural information.

\subsection*{An Interpretation of PSRL}

Corresponding to the true underlying MDP $\mathcal{M}^{True}$ there are $M^N$ stationary deterministic policies. The regret can be minimized if we knew the average reward optimal policy $\pi^*$ of $\mathcal{M}^{True}$. In this case, we would just follow this policy till the end of horizon. However, we do not know the transition probabilities ${P}^{True}$ and rewards ${R}^{True}$ of $\mathcal{M}^{True}$ and thus have to minimize the cumulative regret using the samples. This leads us to reinterpret our regret minimization problem in the famous \textit{multi-arm bandit} problem setting with $M^N$ arms. In this interpretation, each arm corresponds to a policy $\pi$ of $\mathcal{M}^{True}$.  Associated with each arm, we thus have a transition probability matrix ${P}^{True,\pi}$, a reward vector ${R}^{True,\pi}$ and an average reward $\rho(\pi)$. Let $d^{\pi}$ correspond to the stationary distribution vector of the Markov chain induced by  this policy.  The average reward is then $\rho(\pi)=\langle d^{\pi},{R}^{True,\pi} \rangle$ (where $\langle \cdot,\cdot \rangle$ represents inner-product). 

Again, as in the initial description of PSRL, we act in episodes. 
Let us assume a belief on the transition probability matrix ${P}^{True,\pi}$ of the Markov chain under policy $\pi$ and reward vector ${R}^{True,\pi}$ for each policy $\pi$. At the beginning of each episode, let us sample an instance $(P^{\pi},R^{\pi})$ for each policy $\pi$ from the corresponding belief distribution. We can then compute the average reward value for each policy (which may not be equal to $\rho(\pi)$ since we are not using ${P}^{True,\pi}$ and ${R}^{True,\pi}$). We then choose the policy $\pi_{chosen}$ which has the highest computed average reward value to execute in the episode (notice how this is like Thompson sampling in the bandit setting where one picks the arm that has the highest sample value). Using the samples collected during the episode, we can update the belief distribution on ${P}^{True,\pi}$ and ${R}^{True,\pi}$. We repeat this process at the beginning of each episode till the end of the horizon. 

Note that computing average reward estimates for each of the $M^N$ policies and choosing the highest is computationally expensive. This is happening because we are sampling $P^{\pi}$ and $R^{\pi}$ separately for each of the policy.  To alleviate this, the PSRL algorithm can be seen as sampling an MDP  $(P,R)$ and then deriving $\{(P^{\pi},R^{\pi})\}$ for all policies simultaneously. It then determines the policy with the highest estimated average reward using policy iteration (known to converge in a few iterations in practice) without estimating average rewards for each of the $M^N$ policies. 

\subsection*{Policies as arms}

Given this policies-as-arms reinterpretation of PSRL, we can go a step further. We can directly maintain a belief on the average reward $\rho(\pi)$ for each policy. This is precisely what we do in our proposed algorithms in Section \ref{sec:palgos}. Note that this has the advantage that we can reuse ideas from the UCB algorithm and the Thompson sampling algorithm, which are the state-of-the-art algorithms in the multi-armed bandit setting, to our problem. Another particularly attractive feature is that we can work directly with policies rather than the underlying transition probabilities. If we now use the knowledge of the structure of the optimal policy, then we can prune the exponential ($M^N$) number of policies to a number that is polynomial in $M$ and $N$ making the algorithm computationally and statistically attractive. 

One of the \textit{key innovations} that we provide in this paper is \textit{the way the beliefs on the average rewards $\rho(\pi)$ will be updated}. We propose two algorithms, pUCB that is related to the UCB algorithm, and pThompson that is related to the Thompson sampling algorithm. In the vanilla multi-armed bandit setting, the rewards obtained after each episode (which is the same as each round) are independent and identically distributed. On the other hand, the situation is not so clear in our case. How do we decide the length of the episode? Does the episode length influence the bias on the estimates of $\rho(\pi)$? Does the update procedure also influence the bias? We answer these questions using the Renewal-Reward Theorem.

Let $S_1,S_2,...$ be a sequence of positive i.i.d random variables with finite means. The random process $\{X_T\}_{T\geq 0}$ with random variable $X_T = \sup\{n:\sum_{i=1}^{n}S_i \leq T\}$ is called a \emph{renewal process}. Let $W_1,W_2,...$ be a sequence of i.i.d. random variables with finite means. $W_i$ can be interpreted as the reward received during period $S_i$. The process defined by $Y_T = \sum_{i=1}^{X_T}W_i$ is called a \emph{renewal-reward process}. 
\begin{theorem}
The Renewal-Reward theorem (\cite{ross1996stochastic}, Chapter 3, Theorem 3.6.1) states that
\begin{align*}
\lim_{T \rightarrow \infty} \frac{\mathbb{E}[Y_T]}{T} = \frac{\mathbb{E}[W_1]}{\mathbb{E}[S_1]}.
\end{align*}
\end{theorem}
The above theorem states that the expected long run average reward is equal to the expectation of the reward in an \textit{episode} divided by the expectation of the duration of that episode assuming the random variables related to every episode are independent and identically distributed. If we can get unbiased estimates of these two quantities, then we can get an unbiased estimate of the expected long run average reward. Note that an obvious choice such as estimating $\mathbb{E}[\frac{W_1}{S_1}]$ will not lead to correct estimates of the long run average reward. 

Under a given policy $\pi$, by fixing the return times to a fixed state $s_0$ as renewal times, the MDP becomes a renewal reward process. Now we can estimate $\rho(\pi)$ by estimating the reward collected in an episode and the duration of the episode. Note that the reward collected in an episode is a sample for $W_1$ and the duration of the episode is the sample for $S_1$ .
In the algorithms that we propose, we can indeed ensure that the episode lengths are random and that each episode gives an i.i.d. sample (see Section \ref{sec:palgos}, Algorithms \ref{alg:pUCB} and \ref{alg:pThompson}). Note that if we deterministically set episode lengths, then we are introducing bias.

\section{Algorithms}
\label{sec:palgos}

\noindent In our algorithms, we consider policies that has same structure as of the optimal policy.  Let $K$ denote the number of such policies.  These $K$ policies are known in advance.

\subsection{Algorithm \ref{alg:pUCB}: pUCB}
In pUCB algorithm, the set of $K$ policies along with a starting state $s_{start}$, the number of rounds $T$, parameters $\tau$ and $\{\beta(t)\}_{t=1}^{T}$ are provided as input. At the start of the algorithm a random policy is decided to be followed in the episode. After an episode starts, we keep track of the total reward collected (see Line 12 in Algorithm \ref{alg:pUCB} and Line 11 in Algorithm \ref{alg:pThompson}) and the number of time steps elapsed $t'$ before one of the termination conditions is satisfied. The termination conditions are as follows: we end the episode if either the time steps in the episode is equal to $\tau$ or we have reached the starting state $s_{start}$. Note that a key difference from MAB algorithms is that we act in episodes instead of time steps.

\begin{algorithm}[h]
\caption{pUCB algorithm (pUCB)}
\begin{algorithmic}[1]
\STATE \textbf{Input:}  $T, \Pi= \{\pi_k:\;k=1,\hdots,K\}, \{\beta(t)\}_{t=1}^{T}, s_{start}, \tau$
\STATE \textbf{Output:} $CR_{T}$ (Cumulative reward in $T$ rounds)
\FOR {$k=1$ to $K$} 
\STATE $\hat{\rho}(k) = 1,~n(k) = 0$
\STATE $R_{arm}(k) = 0$ (Running sum of rewards for policy $\pi_k$)
\STATE $T_{arm}(k) = 0$ (Running sum of number of rounds for $\pi_k$)
\ENDFOR \\
\STATE $s = s_{start}, ~t' = 0, r = 0$
\STATE $k \sim \textrm{rand}(1,K)$
\FOR {$t=1$ to $T$} 
\IF{ ($t \neq 1$ \textbf{and} $s=s_{start}$) \textbf{or} $t' \geq \tau$}
\STATE $R_{arm}(k) = R_{arm}(k) + r$
\STATE $T_{arm}(k) = T_{arm}(k) + t'$
\STATE $\hat{\rho}(k) = \frac{R_{arm}(k)}{T_{arm}(k)}$
\STATE $c(k) = \beta(t)\sqrt{\frac{2\log t}{n(k)}}$
\STATE $n(k) = n(k) + 1$
\STATE $k = \arg\max_{j=1,...,K}\Big\{ \hat{\rho}(j) + c(j) \Big\}$
\STATE $t' = 0, ~r=0$
\ENDIF \\
\STATE $s_{next} \sim P^{True}(\cdot \mid s,\pi_{k}(s) ),~r = R^{True}(s,a,s_{next})$
\STATE $t' = t' + 1,~s = s_{next}$
\ENDFOR
\STATE $CR_T = \sum_{j=1}^{K}R_{arm}(j)$ 
\end{algorithmic}
\label{alg:pUCB}
\end{algorithm}

We maintain an estimate of the long-run average reward obtained under each policy $\pi_k$ as $\hat{\rho}(k)$. At the end of an episode, we update $\hat{\rho}(k)$ using $r$ and $t'$ in a careful way as shown in line numbers 15-17 in Algorithm \ref{alg:pUCB}. This is justified using the renewal-reward theorem as discussed in Section \ref{sec:prelim}. In the next episode, we follow the policy that has the highest value, of the sum of the average reward estimate $\hat{\rho}(k)$ and the confidence bonus $\beta(t)\sqrt{\frac{2\log t}{n(k)}}$. Here $n(k)$ is used to track the count of the number of times policy $k$ has been picked by round $t$. The algorithm is detailed in Algorithm \ref{alg:pUCB}. 

The sequence $\{\beta(t)\}_{t=1}^{T}$ is an input to the algorithm that determines the exploration-exploitation trade-off as a function of time. If $\beta(t)$ is set to a constant (typically value $1$), then the decision rule is the same as UCB1 of \cite{Auer02}. Note that appropriately choosing this sequence can further decrease regret. Parameter $\tau$ should be set to $\infty$ to ensure our estimates $\hat{\rho(k)}$ remain unbiased. When $\tau = \infty$, we can only switch between policies at the end of recurrent cycles. Mean recurrence times may potentially be large and are dependent on the unknown transition probabilities and the current policy being used.  If they are indeed large, then $\tau$ can let us switch between policies at the expense of getting biased estimates of $\rho(\pi)$. On the other hand, if they are small relative to $\tau$, then setting $\tau$ to a finite value does not affect estimation quality.

\subsection{Algorithm \ref{alg:pThompson}: pThompson}

\begin{algorithm}[h]
\caption{pThompson algorithm (pThompson) }
\begin{algorithmic}[1]

\STATE \textbf{Input:}  $T, \Pi= \{\pi_k:\;k=1,\hdots,K\}, s_{start}, \tau$
\STATE \textbf{Output:} $CR_{T}$ (Cumulative reward in $T$ rounds)
\FOR {$k=1$ to $K$} 
\STATE $CR_{T} = 0,~S(k) = 0, ~F(K)=0$
\ENDFOR \\
\STATE $s = s_{start},~t' = 0,~r=0$
\STATE $k \sim \textrm{rand}(1,K)$

\FOR {$t=1$ to $T$} 
\IF{ ($t \neq 1$ \textbf{and} $s=s_{start}$) \textbf{or} $t' \geq \tau$}
\STATE $CR_T = CR_T + r$
\STATE $S(k)=S(k)+r$
\STATE  $F(k)=F(k)+t' - r$
\FOR {$k=1$ to $K$} 
\STATE $\theta(k) \sim Beta(S(k),F(k))$
\ENDFOR \\
\STATE $k = \arg\max_{j=1,...,K} \Big\{ \theta(j) \Big\}$
\STATE $t' = 0, r=0$
\ENDIF \\
\STATE $s_{next} \sim P^{True}(\cdot \mid s,\pi_{k}(s) ),~ r = R^{True}(s,a,s_{next})$
\STATE $t' = t' + 1,~s = s_{next}$
\ENDFOR 
\end{algorithmic}
\label{alg:pThompson}
\end{algorithm}

Algorithm \ref{alg:pThompson} illustrates our second algorithm pThompson. Its structure, inputs and output are similar to the pUCB algorithm described above and we will focus on the differences here. In terms of inputs, pThompson does not have the sequence $\{\beta(t)\}_{t=1}^{T}$ as one of its inputs.

The initialization for pThompson is similar to that of pUCB except that pThompson maintains a different set of internal estimates. In particular, for each policy $\pi_k$, it maintains two estimates $S(k)$ and $F(k)$. These two estimates parameterize a Beta distribution that encodes our beliefs on the average cost reward of policy $\pi_k$. During each episode, we keep track of the total reward collected $r$ and the number of rounds $t'$ elapsed before either of the termination conditions are met. The cumulative reward for the episode $r$ is added to the running estimate $S(k)$ of the current policy $k$ and an update of the form $t - r$ is done to $F(k)$. This update step is critical and ensure that the mean of the beta distribution is an unbiased estimate of average reward $\rho(k)$. This is different from the update step in Thompson sampling. Note that our updates also rely on conjugacy properties \cite{Agrawal11} (similar to Beta-Binomial conjugacy used in Thompson sampling). Then for new policy selection, we draw a realization for each of the $K$ Beta distributions and pick that policy whose realization value is the highest.

It is important to note that although we mention the use of structural information, such information is only used to define the inputs to the algorithms proposed. Our algorithms work for many different classes of MDPs (including MDPs where there is no structure leading to exponential number of policies, although this may render our algorithms non-competitive). Structural properties are application/MDP specific, and we do not outline how to derive structural properties here.  Obtaining structural properties is highly non-trivial and mathematical and the process of extracting such structural information and using it is not automatic. See references to several applications in Section \ref{introduction} for more details on deriving structural properties, which typically reduce the policy search space dramatically. Indeed, a complete characterization of the class of MDPs based on structural properties of the optimal policy can be a separate research thread in itself. In this sense, pUCB and pThompson are structure agnostic RL regret minimization algorithms, which when coupled with structure restricted policy spaces can compete with the state-of-the-art.

\subsection*{Comparison with UCRL and PSRL}
The UCRL and PSRL algorithms maintain $O(M^2N)$ estimates internally. This is significantly higher compared to our algorithms, that typically maintain $O(M)$ estimates, as seen in the settings of Section \ref{sec:numerical}. Further, PSRL as described in \cite{Osband13} needs to reset to the starting state after the end of the each episode (we take this into account carefully in our experiments). Another important aspect of our algorithms is that they do not incur high sampling costs that are inherently necessary for PSRL. In PSRL, we have to sample $O(M^2N)$ transition probability values and reward values from a belief that we maintain over them. Without using conjugacy, belief updates also become expensive to compute.

\subsection{Algorithm \ref{alg:warmPSRL}: warmPSRL}
For both algorithms, pUCB and the pThompson, we can simultaneously estimate, at any round $t \in \{0,T\}$, the unknown transition probabilities and reward values. For estimating transition probabilities, we keep track of the counts of state transitions for every action. To estimate rewards, we take empirical averages of rewards observed for each state-action pair.  This is beneficial in settings where the policies-as-arms based algorithms such as ours can be used in conjunction with algorithms such as PSRL to further improve on the cumulative rewards collected. 
These estimates can be used to warmstart PSRL.
This leads to the algorithm \textbf{warmPSRL} that we have listed in Algorithm \ref{alg:warmPSRL}. In this algorithm, we provide an additional input $T_{switch}$ that is chosen depending on problem instance. For the initial $T_{switch}$ rounds, we run 
modified versions of pUCB or pThompson (pUCB-Extended and pThompson-Extended respectively) or any other bandit algorithm, where 
we empirically estimate transition probabilities and rewards in parallel. For $T-T_{switch}$, we run PSRL algorithm with the estimates computed by our algorithms as the initialization values. Finally, note that the warmPSRL is a combination of model free and model based methods.

\begin{algorithm}
\caption{PSRL warmstarted with pUCB or pThompson }

\begin{algorithmic}[1]
\STATE \textbf{Input:}  $T, \Pi= \{\pi_k:\;k=1,\hdots,K\}, s_{start}, \tau, T_{switch}$, $\{\beta(t)\}_{t=1}^{T_{switch}}$,\\
\quad\quad\quad Alg = pUCB-Extended or pThompson-Extended.
\STATE \textbf{Output:} $CR_T$ (Cumulative reward in $T$ rounds)
\STATE \textbf{Execution}	      
\STATE $(CR_{1},\hat{P},\hat{R})$ = Alg($T_{switch}, \Pi, \{\beta(t)\}, s_{start}, \tau$) \/\/
\STATE $CR_2$ = PSRL($T-T_{switch},\tau,\hat{P},\hat{R},s_{start}$)
\STATE $CR_T = CR_1 + CR_2$
\end{algorithmic}
\label{alg:warmPSRL}
\end{algorithm}

\subsection*{Bounds on regret}

Here we provide a brief sketch of the regret analysis for pUCB and pThompson algorithms.

For the following analysis, we will assume $\tau=\infty$. 
By the unichain assumption on MDP, Markov chains under all policies have a single recurrent class.  Further, irreducibility implies that the mean recurrence time starting from any state will be finite.
Let us consider the recurrent times to state $s_{start}$ under all policies. Let $\eta_{max}= \underset{\pi_k, ~k \in \{1,2,\ldots,K\}}{\max} E_{\pi_k}[s_0 \rightarrow s_0]$, where $E_{\pi_k}[s_0 \rightarrow s_0]$ denotes the mean recurrent time to state $s_{start}$ under policy $\pi_k$. 

Note that $\eta_{max}$ is an upperbound on mean recurrent times under all policies. In a horizon length $T$, on an average, we can have at most $\frac{T}{\eta_{max}}$ \emph{pulls} of the policies. Now we have a multi-arm bandit setting with horizon length $\frac{T}{\eta_{max}}$ instead of the original horizon length $T$ in the MDP setting. From this reduction, it is clear one can use the regret bounds of UCB as well as the Thompson sampling algorithm with the total number of pulls being $\frac{T}{\eta_{max}}$. This leads to a cumulative regret of $O(\log(\frac{T}{\eta_{max}}))$, i.e., logarithmic pulls of sub-optimal arms. A rigorous analysis of this result involves analyzing the average number of pulls of the \emph{arms} $N_{\pi_k}(t)$ by time $t$ and utilizing the renewal reward theorem to establish the convergence of average of i.i.d bounded rewards to the average reward of policy $\pi_k$. Note that we can use Hoeffding's inequality for Markov chains to get the rate of convergence. The proofs of pUCB and pThompson algorithms then will respectively follow along the lines of UCB and Thompson sampling algorithms giving us the needed regret bounds. 

\section{Numerical Results}
\label{sec:numerical}

\noindent Below we describe two MDP settings where there is structure in the corresponding optimal policies. Knowing this information simplifies the search for the policy that minimizes regret.

\subsection{The slow server problem}
Consider a service system comprising of $K$ servers and a single queue where the arriving customers wait before obtaining service. The customers arrive according to a Poisson process with rate $\lambda$ and each customer is 
required to obtain service at one of the parallel servers. Let us assume that the service requirement of each arriving customer is unity and that the time taken by server~$k$ to serve a customer is a random variable which has an exponential distribution with rate $\mu_k$ for $k = 1,\ldots,K.$ This service system reflects many practical scenarios (including call center operations, web server load balancing and others).

This problem is well studied \cite{Lin84,Walrand84,Koole95} in the setting where  $K = 2$, $\mu_1 > \mu_2$ and all parameters are known. In particular, it has been shown that in an optimal admission control policy, the faster server should always be occupied when there are customers waiting for service in the queue. Further, the optimal policy is of the threshold type, which means that the slower of the two servers must be utilized only when the number of customers in the queue exceeds a threshold. While the optimal policy is of the threshold type, no known methods seem to exist that characterize the threshold in terms of the parameters.

\begin{figure}
\centering
\includegraphics[width=.7\columnwidth]{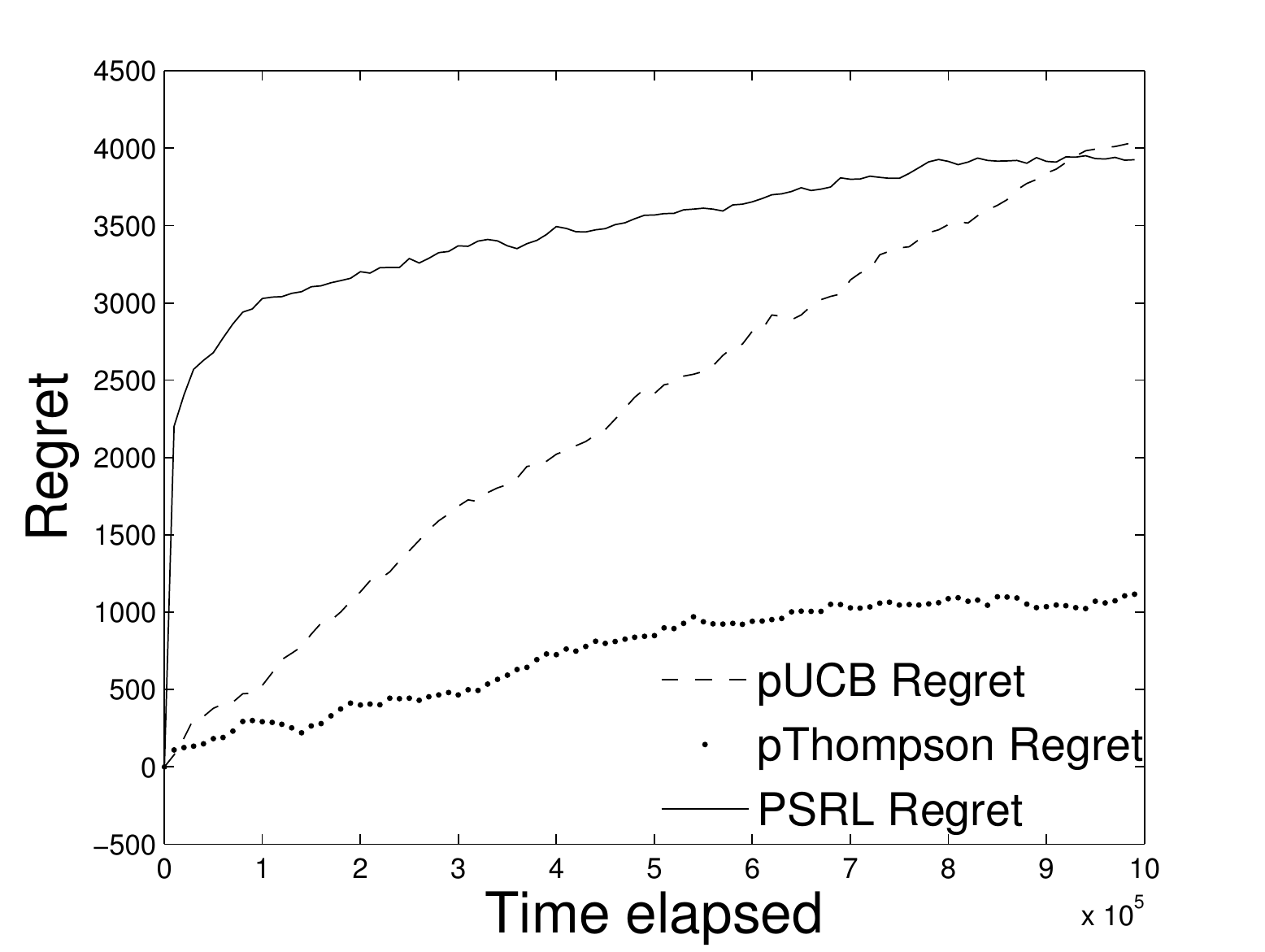}
\caption{Regret as a function of number of rounds for the slow server problem.\label{fig:queueing}}
\end{figure}

For our experiment, we chose $\lambda = \frac{12}{31}$, $\mu_1 = \frac{18}{31}$ and $\mu_2 = \frac{1}{31}$. The buffer length of the queue was chosen to be $20$ (thus leading to a number of states equal to $80$). We ran 10 Monte Carlo simulations and the resulting regret achieved by the proposed algorithms is shown in Figure \ref{fig:queueing}. We compare our algorithms with PSRL. UCRL was not used for comparison because incorporating state dependent action spaces into the algorithm is difficult. On the other hand, for PSRL, we were able to do this by suitably modifying the Dirichlet distribution \cite{Osband13}. We ran each simulation for $10^6$ rounds. The starting state corresponds to the state where the queue is empty and both the servers are free. The parameter $\tau$ was set to $\infty$ for pUCB and pThompson. Further, $\beta(t)$ was set to $1$ for pUCB. 

As shown in the plot, pUCB and pThompson perform much better than the state-of-the-art algorithm PSRL from the very outset. This clearly illustrates the advantage of using knowledge of the optimal policy structure.

\subsection{The machine replacement problem}
The setting for this problem is adapted from \cite{Bert05} (Example 1.1.3, page 8) and 
has various applications in inventory management. Consider the problem of operating a machine efficiently. The machine can be in one of $n$ possible states ($S = \{1,2,\ldots,n\}$). Let state $1$ correspond to the machine being in perfect condition and each subsequent state corresponding to increasingly deteriorated condition of the machine.Let there be an average cost $g(i)$ for operating the machine for one time period when it is in state $i$. Because of the increasing failure probability, we can assume that $g(1) \leq g(2) \leq \cdots \leq g(n)$. We can take two actions in each state: continue operating the machine without maintenance ($C$) or perform maintenance ($PM$).  Once maintenance has been performed, the machine is guaranteed to remain in state $1$ for one time period.The cost that we incur for maintenance is thus $R+g(1)$ ($R$ for repairing and $g(1)$ because the machine is now functioning in state $1$).

 Let $P = [[p_{ij}(a)]]$, $i,j\in S$, $a\in \{C,PM\}$ denote the transition probability matrix, with the following properties: (a) $p_{i1}(PM)=1$, (b) $p_{ij}(PM)=0,$ $\forall j\neq 1$, (c) 
 $p_{ij}(C) =0,$ $\forall j<i$, and (d) $p_{ij}(C) \leq p_{(i+1)j}(C),$ $\forall j>i$. Intuitively, when the machine is operated in state $j$, its well-being will deteriorate to another state $i \geq j$ after the current time period.
 
For this application and many others based on it, the optimal policy is a threshold policy if our objective is to minimize the average cost of using the machine. That is, we should perform maintenance if and only if the state of the machine $i \geq i^*$, where $i^*$ is a certain threshold state. This threshold state can be identified if we know the precise transition probability values.

\begin{figure}
\centering
\includegraphics[width=.7\columnwidth]{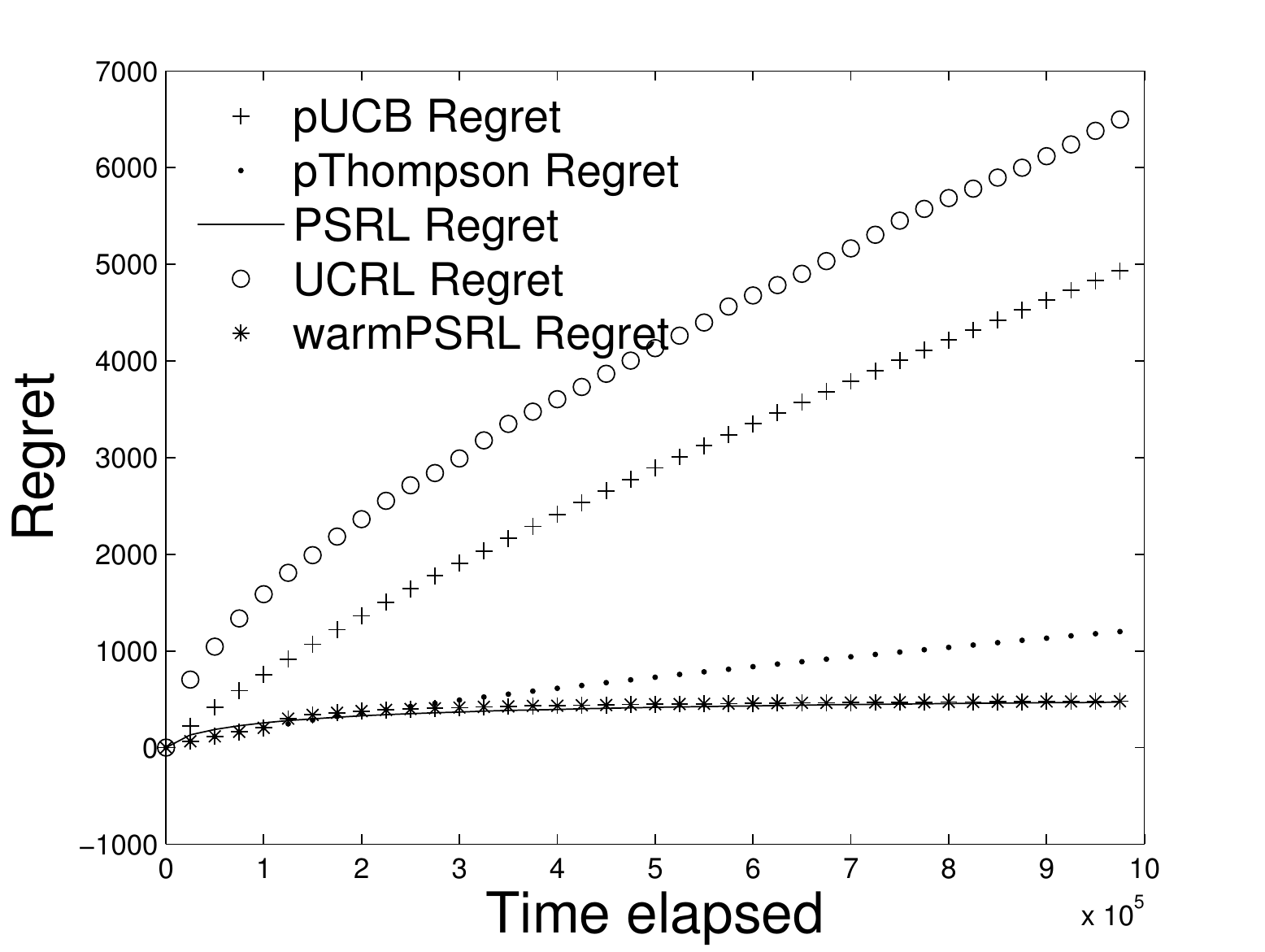}
\caption{Regret as a function of number of rounds for the problem of machine replacement.\label{fig:machineReplacementCombined}}
\end{figure}

We chose the following experimental configuration. The number of states was chosen to be $100$. We ran $10$ Monte Carlo simulations and the resulting regret achieved by the proposed algorithms is shown in Figure \ref{fig:machineReplacementCombined}. The true transition probability values were generated randomly (taking into account the constraints relating these values) and were kept fixed for each simulation run. The algorithms that we compare to are PSRL and UCRL. We ran each simulation for $10^6$ rounds. The starting state corresponds to the state where the machine is in perfect condition. The parameter $\tau$ was set to $\infty$ for pUCB and pThompson. Further, $\beta(t)$ was set to $1$ for pUCB. In warmPSRL, we used pThompson for $10^5$ rounds, estimated $(P,R)$ and then switched to PSRL with the estimated (P,R) as the starting values for the remaining rounds. Appropriate best values were chosen for PSRL and UCRL parameters as well. 
 
As we can see from the plot, PSRL, warmPSRL and pThompson are very close in terms of performance. In fact, pThompson is better than PSRL for the first $10^5$ rounds. Around this point, PSRL has learned accurate enough estimates of transition probabilities and reward values that it seems to collect relatively slightly more reward per round. Thus, if fast initial `learning' is needed or if lesser rate of regret is desired in fewer number of rounds, pThompson algorithm can be chosen since it outperforms state-of-the-art PSRL (as shown in the plot). The regret of warmPSRL is very close to that of PSRL overall and better in the initial rounds. Among the remaining two algorithms, pUCB performs better than UCRL although neither of them are as good as pThompson, warmPSRL and PSRL. 
Finally, note that our algorithms ran much faster than PSRL and UCRL (also true in the preceding experiment, although running times are not reported for both).

\section{Concluding Remarks}
 \label{sec:summary}
 
\noindent We have built on the well established, easy to use UCB and Thompson sampling algorithms to provide competitive regret minimizing RL algorithms (making novel modifications for getting unbiased estimated of the long run average reward $\rho(k)$ and the number of rounds that each policy $\pi_k$ needs to be applied). Such direct use of structural information is not easily possible for the current state of the art algorithms (PSRL and UCRL). The algorithms presented are simple, competitive and many times better than state-of-the-art methods for cumulative regret minimization in RL settings. They have significant advantage in terms of computation and sampling costs.

\clearpage

\bibliographystyle{myIEEEtran}
\bibliography{ref}
\end{document}